\documentclass[letterpaper]{article} 
\usepackage[]{aaai23}  
\usepackage{times}  
\usepackage{helvet}  
\usepackage{courier}  
\usepackage[hyphens]{url}  
\usepackage{graphicx} 
\usepackage{natbib}  
\usepackage{caption} 
\usepackage{algorithm}

\usepackage{amsmath}
\usepackage{algorithmicx}
\usepackage{amssymb}
\usepackage{graphicx}
\usepackage{multirow}
\usepackage{booktabs}
\usepackage{dsfont}
\usepackage{bm}
\usepackage{algpseudocode}
\usepackage{subfigure}
\usepackage{color}
\usepackage{fdsymbol}
\usepackage{CJKutf8}

\usepackage{newfloat}
\usepackage{listings}
\DeclareCaptionStyle{ruled}{labelfont=normalfont,labelsep=colon,strut=off} 
\floatstyle{ruled}
\newfloat{listing}{tb}{lst}{}
\floatname{listing}{Listing}
\title{Zero-Shot Rumor Detection with Propagation Structure via Prompt Learning}
\author{
    Hongzhan Lin$^1$\equalcontrib,
    Pengyao Yi$^2$\equalcontrib,
    Jing Ma$^1$\thanks{Corresponding authors.},
    Haiyun Jiang$^{3\dagger}$,\\
    Ziyang Luo$^1$,
    Shuming Shi$^4$,
    Ruifang Liu$^2$
}
\affiliations{
    \textsuperscript{1}Hong Kong Baptist University\\
    \textsuperscript{2}Beijing University of Posts and Telecommunications\\
    \textsuperscript{3}Fudan University \textsuperscript{4}Tsinghua University\\
    \{cshzlin, majing, cszyluo\}@comp.hkbu.edu.hk, jianglu@fudan.edu.cn, \{yi.py, lrf\}@bupt.edu.cn

}

\usepackage{bibentry}

\begin{document}

\maketitle

\begin{abstract}
The spread of rumors along with breaking events seriously hinders the truth in the era of social media. Previous studies reveal that due to the lack of annotated resources, rumors presented in minority languages are hard to be detected. Furthermore, the unforeseen breaking events not involved in yesterday's news exacerbate the scarcity of data resources. In this work, we propose a novel zero-shot framework based on prompt learning to detect rumors falling in different domains or presented in different languages. More specifically, we firstly represent rumor circulated on social media as diverse propagation threads, then design a hierarchical prompt encoding mechanism to learn language-agnostic contextual representations for both prompts and rumor data. To further enhance domain adaptation, we model the domain-invariant structural features from the propagation threads, to incorporate structural position representations of influential community response. In addition, a new virtual response augmentation method is used to improve model training. Extensive experiments conducted on three real-world datasets demonstrate that our proposed model achieves much better performance than state-of-the-art methods and exhibits a superior capacity for detecting rumors at early stages.
\end{abstract}  

\section{Introduction}
The spread of rumors emerging along with breaking news is a global phenomenon, which can cause critical consequences for social network users in different lingual contexts. 
For example,  during the unprecedented COVID-19 pandemic, a false rumor claiming that ``the vaccine has a chip in it which will control your mind"\footnote{\url{https://www.bbc.com/news/55768656}} released by a Muslim cleric, went viral on Facebook and Twitter in different languages. Such misleading claims about vaccines are being shared widely in many countries, which confuse the public and undermine their enthusiasm for vaccination. Due to the barriers of domain and language, even human fact-checkers are poor judges of such rumors. Therefore, it's imperative to develop automatic approaches for rumor detection spread in different languages amid unforeseen breaking events.

\begin{figure}[t]
\centering
\scalebox{0.40}{\includegraphics[width=20cm]{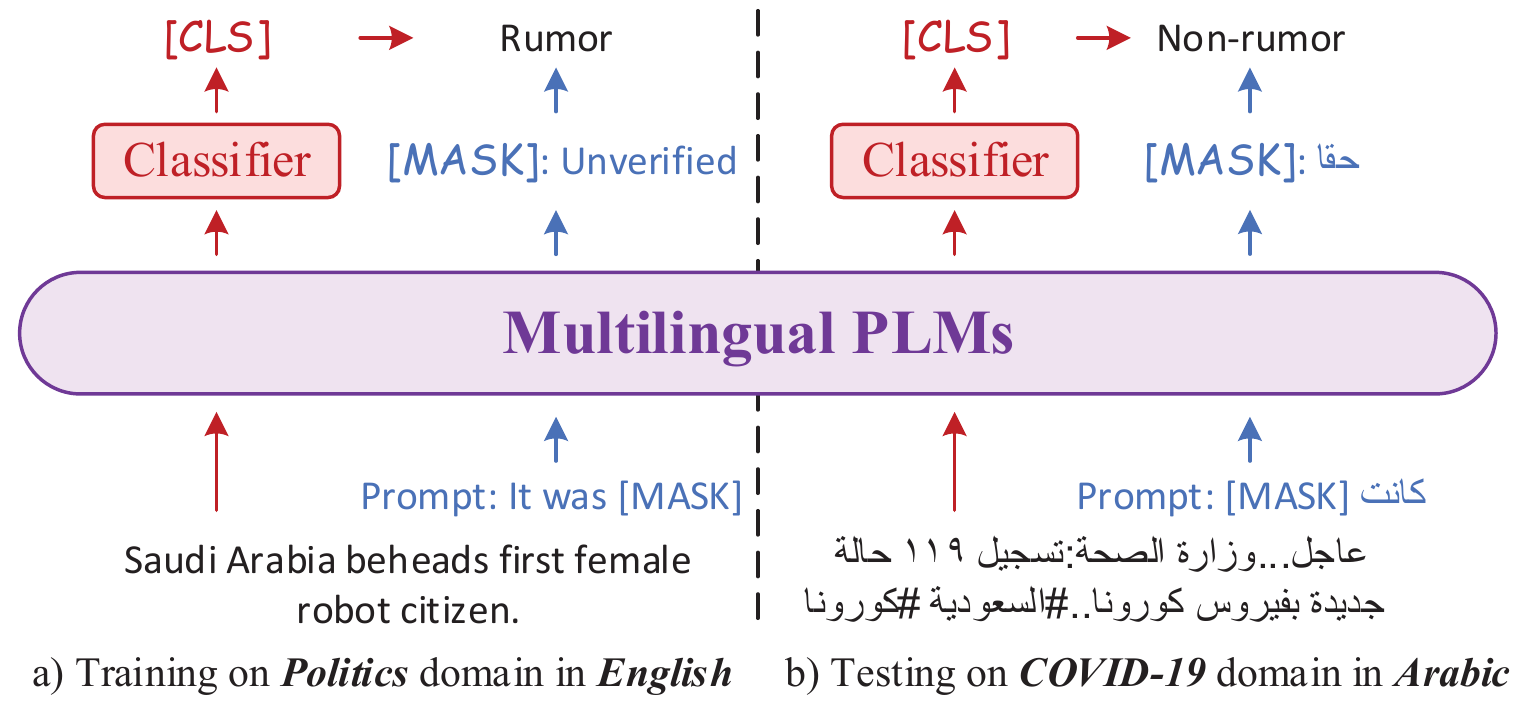}}
\caption{Illustration between the task-specific fine-tuning and the prompt learning paradigms for solving ZRD task.}
\label{fig:task_illu}
\vspace{-0.5cm}
\end{figure}

Social psychology literature defines a rumor as a story or a statement whose truth value is unverified or deliberately false~\cite{allport1947psychology}. In this study, we focus on detecting rumors on social media, instead of ``fake news" strictly defined as a news article published by a news outlet that is verifiably false~\cite{yang2022coarse}. 
State-of-the-art techniques using deep neural networks (DNNs)~\cite{bian2020rumor, lin2021rumor, rao2021stanker} have promoted the development of rumor detection, but they are all data-driven models that require extensive annotated data for model training. Most corpora are open-domain and presented in English, 
which makes them not scalable to emerging events in new languages where only few/no labeled data is available. 
Zero-shot rumor detection task (ZRD) aims to adapt knowledge learned in the source rumor data to the target data without labeled training samples in the target language and domain, as shown in Figure~\ref{fig:task_illu}. Previous related studies~\cite{du2021cross,tian2021rumour} directly utilize pre-trained language models (PLMs)~\cite{devlin2019bert} to fine-tune on ZRD task. 
However, they just formulated the zero-shot rumor detection as a cross-lingual text classification problem and detected the single claim post with a heavy task-specific fine-tuning stage, which makes it deviate from the pre-training target on masked language modeling, even ignoring the domain-invariant interaction of user opinions during the diffusion of rumors. 
More recently, \citet{lin2022detect} propose a contrastive learning framework to detect rumors from different languages and domains, where a small number of target annotation is required. However, it is prone to be poor at 
emerging events propagated in minority languages without any expertise annotation, especially in some underdeveloped countries and regions. For breaking events with scarce annotated data in different languages, the study on the zero-shot regimes is more urgent and practical for rumor detection on social media.

In this paper, we focus on exploring efficient prompting with language and domain transfer for zero-shot rumor detection. We assume there are no accessible annotations in the target language and domain, so prompt learning mechanisms~\cite{zhao2021discrete} based on existing multilingual PLMs can be utilized. However, the standard prompt learning paradigm adopts discrete or soft prompts, where the discrete prompt requires experts of native speakers to design rumor-related templates/rules for different languages, and the soft prompt uses optimized token representations trained on a large dataset. Unlike the standard prompt-tuning paradigm, we propose to decouple shared semantic information from the syntactic bias in specific languages based on multilingual PLMs, which could enhance the semantic interaction between the prompt and rumor data. Besides, as the diffusion of rumors generally follows spatial and temporal relations that provide valuable clues on how a claim is transmitted irrespective of specific domains~\cite{zubiaga2018detection}, we aim to develop a novel prompt learning mechanism to take such social context into consideration.

To this end, we propose a zero-shot Response-aware Prompt Learning (RPL) framework to detect cross-lingual and cross-domain rumors on social media. 
More specifically, we firstly rank responsive posts toward the claim to represent diverse propagation threads. Then
a hierarchical prompt encoding mechanism is proposed based on multilingual PLMs, which alleviates the effort of prompt designing for different languages.
On the other hand, as the propagation structure contains domain-invariant features on how a claim is responded to by users over time, we model the absolute and relative propagation position to capture the latent structure of the propagation thread for better domain adaptation.
To further improve the zero-shot model training, we incorporate a new virtual response augmentation mechanism into the prompt learning framework. As there is no public benchmark available for detecting rumors in low-resource languages with propagation threads in tweets, we collected a new rumor dataset corresponding to COVID-19 from Twitter in Cantonese and Arabic languages. 
Extensive experiments conducted on three real-world rumor datasets corresponding to COVID-19 confirm that (1) our model yields outstanding performance for detecting zero-shot rumors over the state-of-the-art baselines with a large margin; and (2) our method performs particularly well on early rumor detection which is crucial for timely intervention and debunking especially for breaking events. 

\section{Related Work}
Pioneer studies for automatic rumor detection focused on learning a supervised classifier utilizing features crafted from post contents, user profiles, and propagation patterns~\cite{castillo2011information, yang2012automatic, liu2015real}. Subsequent studies then proposed new features such as those representing rumor diffusion and cascades \cite{kwon2013prominent, friggeri2014rumor, hannak2014get}. \citet{zhao2015enquiring} alleviated the engineering effort by using a set of regular expressions to find questing and denying tweets. Deep neural networks such as recurrent neural networks~\cite{ma2016detecting}, convolutional neural networks~\cite{yu2017convolutional}, and attention mechanism~\cite{guo2018rumor} were then employed to learn the features from the stream of social media posts. To extract useful clues jointly from content semantics and propagation structures, some approaches proposed kernel-learning models~\cite{wu2015false, ma2017detect}, tree-structured recursive neural networks (RvNN)~\cite{ma2018rumor}, self-attention models (PLAN~\cite{khoo2020interpretable}, STANKER~\cite{rao2021stanker}) and graph neural networks (BiGCN)~\cite{bian2020rumor} have been exploited to encode conversation threads for higher-level representations.


Recently, zero-shot transfer learning techniques are applied on PLMs to detect fake news~\cite{du2021cross, schwarz2020emet, de2021transformer} by downstream task-specific fine-tuning methods. \citet{tian2021rumour} utilized PLMs and a self-training loop to adapt the model from the source language to the target language in multi-step iteration. However, these approaches only consider cross-lingual text classification and face problems such as task mismatch between pre-training and fine-tuning, and ignore the domain-invariant propagation patterns from community response. Considering that rumors can be domain-specific and/or presented in different languages, \citet{lin2022detect} first introduced supervised contrastive learning for few-shot rumor detection based on propagation structure. However, their few-shot paradigm still relies on a small number of target data for training, which cannot perform well in detecting more minority-language rumor data without any expertise annotation in case of emerging topics.

Prompt learning converts downstream tasks to language modeling tasks via textual prompts, which is found more effective to use PLMs than typical fine-tuning on specific tasks~\cite{brown2020language,liu2021pre}. In recent years, prompt learning has achieved great success in a variety of NLP tasks, such as text classification~\cite{min2021noisy}, semantic parsing~\cite{schucher2021power}, text generation~\cite{li2021prefix}, sentiment classification~\cite{seoh2021open} and dialog state tracking~\cite{lee2021dialogue}, etc. Despite the flourish of the research in prompting methods, there is only limited attention being put on the low-resource rumor detection task~\cite{lin2022detect}. Different from a few previous multilingual work~\cite{zhao2021discrete, winata2021language, lin2021few} on either discrete or soft~\cite{liu2021p, lester2021power} prompts, in this paper, we tune the level-grained models’ parameters for language-agnostic rumor prompts, which further attends to user interactions from community response for zero-shot rumor detection task.

\begin{figure*}[t]
\centering
\scalebox{0.73648}{\includegraphics[width=20cm]{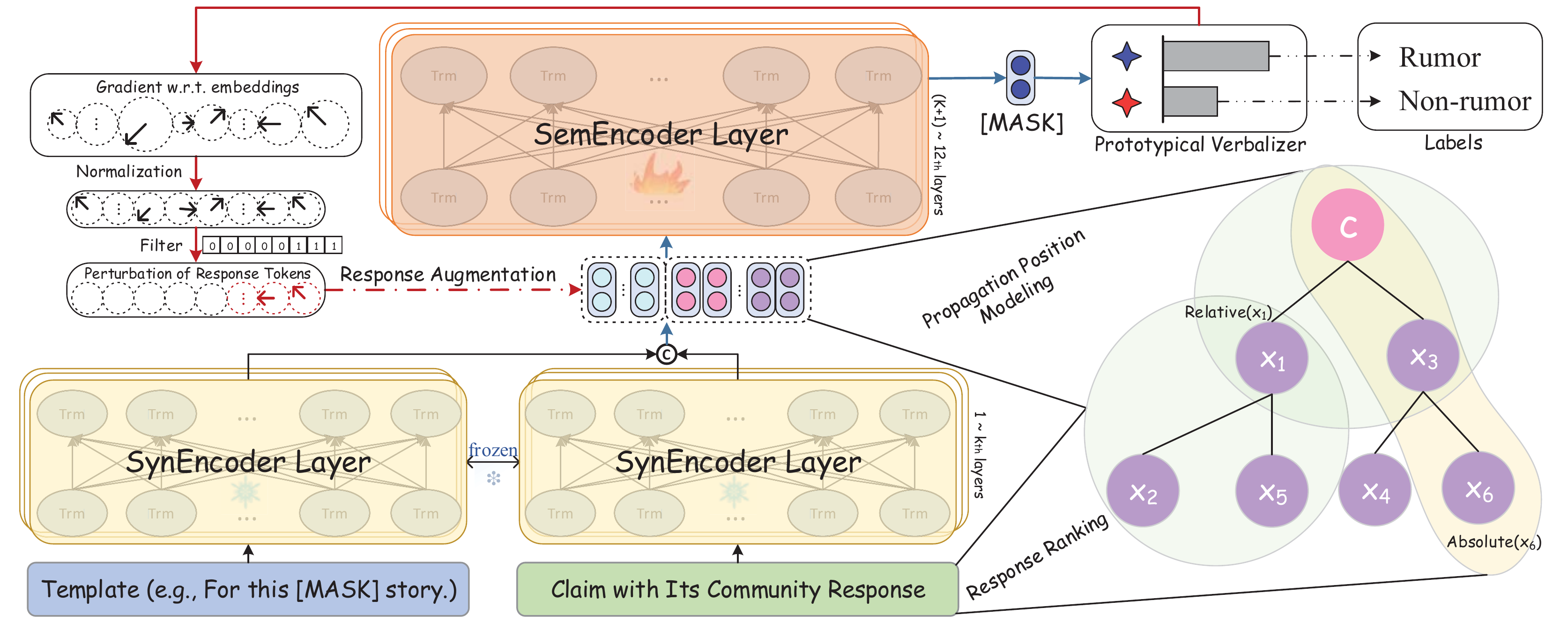}}
\caption{The overall architecture of our proposed method. For the source training data, we first obtain intermediate syntax-independent embeddings after the SynEncoder, then tune the SemEncoder with the prototypical verbalizer paradigm. For target test data, the similarity score between the output states of \emph{[MASK]} and prototypes would be used to detect rumors.}
\label{fig:method}
\vspace{-0.5cm}
\end{figure*}

\section{Problem Statement and Background}
In this work, we define the zero-shot rumor detection task as: given a dataset as source, classify each event in the target dataset as a rumor or not, where the source and target data are from different languages and domains. Specifically, we define a source dataset for training as a set of events $\mathcal{D}_s = \{C_1^s, C_2^s, \cdots, C_{M}^s\}$, where $M$ is the number of source events. Each event $C^s=(y,c,\mathcal{T}(c))$ is a triplet representing a given claim $c$ which is associated with a veracity label $y \in \{\text{rumor}, \text{non-rumor}\}$, and ideally all its relevant responsive microblog post in chronological order, i.e., $\mathcal{T}(c) = \left[ x_{1}^s,x_{2}^s,\cdots,x_{m}^{s}\right]$, where $m$ is the number of responsive posts in the conversation thread. We consider the target dataset with a different language and domain from the source dataset for testing $\mathcal{D}_t = \{C_1^t, C_2^t, \cdots, C_{N}^t\}$, where $N$ is the number of target events and each $C^t=(c',\mathcal{T}(c'))$ shares the similar structure as that of the source.

This task could be formulated as a supervised classification problem that trains a language/domain-agnostic classifier $f(\cdot)$ transferring the features learned from source datasets to that of the target events, that is, $f(C^t| \mathcal{D}_s) \rightarrow y$. 

In this work, we convert the rumor detection as a cloze-style masked language modeling problem. For example, given a cloze-style template $p$ (e.g., ``\emph{For this [MASK] story.}''
as the prompt, spliced with the claim $c$ into $\hat{c}$, the standard prompt learning leverages PLMs to obtain the hidden state for the $\emph{[MASK]}$ token, to infer the rumor-indicative words to fill in $\emph{[MASK]}$.
The probability of label $y$ is:
{
\begin{equation}\label{eq:1}
\begin{aligned}
    \mathcal{P}(y|\hat{c})=g(\mathcal{P}(\emph{[MASK]}=v|\hat{c})|v \in \mathcal{V}_y)
\end{aligned}
\end{equation}
} where $\mathcal{V}$ is a set of rumor-related label words, $\mathcal{V}_y$ is the subset of $\mathcal{V}$ corresponding to $y$ and $g(\cdot)$ is a manual verbalizer to transform the probability of label words into that of the label. In this way, we could map predicted words for $\emph{[MASK]}$ into the veracity label to make a decision on the claim.

\section{Our Approach}
In this section, we introduce our Response-aware Prompt Learning framework for zero-shot rumor detection, in cross-lingual and cross-domain settings. Because the rumor-related prompt design for different languages can be biased and labor-intensive, we propose to learn language-independent prompts including the template and verbalizer. On another hand, as the responsive posts could provide a domain-invariant propagation structure for representation learning, we explore how to fuse such community response into the prompt learning framework. Figure~\ref{fig:method} illustrates an overview of our proposed model, which includes: 
1) Response Ranking, which presents each event as diverse propagation threads following temporal or spatial relations; 2) Hierarchical Prompt Encoding, which is the backbone to learn language-independent interaction between the prompt and the event with the prior knowledge of multilingual PLMs; 3) Propagation Position Modeling, which equips our proposed prompt-based framework with the latent structure of the propagation thread; and 4) Response Augmentation, which adds noise to responsive posts to improve model training for better robustness. 

\subsection{Response Ranking}
To highlight the social context for enhancing the contextual representation learning for the event, we propose to attend over evidential responses. The core idea is to rank all the responses based on diverse propagation threads. 

First, we hypothesize that the attitudes of responsive posts towards the claim will become more inclined as time goes by, thus the responsive posts can be sorted in chronological and inverted order on the time sequence. Specifically, for the chronological order, responsive posts with earlier time stamps are prioritized, i.e., $\mathcal{T}(c) = \left[ x_{1},x_{2},\cdots,x_{m}\right]$, and vice versa for the inverted order on the time sequence, i.e.,  $\mathcal{T}(c) = \left[x_{m},x_{m-1},\cdots,x_{1}\right]$. 

Besides the perspective of time sequence, inspired by~\cite{ma2018rumor,bian2020rumor}, we further represent the propagation thread as a tree structure $\mathcal{T}(c)=\langle \mathcal{G}, \overrightarrow{\mathcal{E}} \rangle$, where $\mathcal{G}$ refers to a set of nodes each representing a responsive post of ${c}$, and $\overrightarrow{\mathcal{E}}$ is a set of directed paths conforming to the responsive relation among the nodes in $\mathcal{G}$. 
We scrutinize the optimal search algorithms on the tree structure to select more evidential posts in depth-first and breadth-first order. Specifically, the depth-first search studies the propagation patterns during information flows from the ancestor to the children nodes while the breadth-first search gives priority to the interaction of user opinions among sibling nodes. Taking the propagation tree in Figure~\ref{fig:method} as an example, the depth-first order of the response ranking would be $\left[ x_{1},x_{2},x_{5}, x_{3}, x_{4}, x_{6}\right]$; for the breadth-first order, it would be $\left[ x_{1},x_{3},x_{2}, x_{4}, x_{5}, x_{6}\right]$.

In this way, concerning perspectives of time sequence or propagation tree, we could investigate the importance of different responses $\mathcal{T}(c)$ on the verdict of a claim. 

\subsection{Hierarchical Prompt Encoding}
Generally, it will lead to bias towards syntax in specific languages if we directly utilize the existing tokens from the vocabulary like expertise words or language-specific slang as the template. To bridge the gap between languages in this task, the template shall not depend on any specific language. Although the soft prompt is a potential way to solve this problem, its trainable tokens require enough target rumor data for training, which is challenging in zero-shot regimes.
To this end, we hope to implicitly disentangle shared semantic information from different languages with language-specific syntactic knowledge, by leveraging the priors of multilingual PLMs. Previous literature~\cite{jawahar2019does, rao2021stanker, huang2022zero} has shown that the lower layers of PLMs can capture syntactic-level features while the upper layers of PLMs model the semantic-level features. 
Therefore, we can present a Hierarchical Prompt Encoding (HPE) mechanism for language-independent representation learning of the template and the event at syntactic and semantic levels. In our approach, we hypothesize that the semantic information could be shared over different languages though the syntax is language-dependent. 

\textbf{SynEncoder Layer.} At the syntactic level, to obtain the intermediate syntax-independent embeddings, we copy and froze the parameters of the lower $k$ layers from multilingual PLMs encoders to encode the template and the event, respectively. Specifically, the original template $p$ is syntactically mapped into a shared vector space:
{
\begin{equation}\label{eq:2}
\begin{aligned}
    {X}_{p} = \operatorname{SynEncoder}\left({p}\right)
\end{aligned}
\end{equation}
} where ${X}_p \in \mathbb{R}^{|p| \times d}$ is the template embeddings and $d$ is the dimension of the output state of SynEncoder.

For an event $C$, as all the responsive posts are presented in the same language and domain as the claim either at the training or testing stages, we could concatenate them in the same frozen SynEncoder to obtain the embeddings of the event:
{
\begin{equation}\label{eq:3}
\begin{aligned}
    {X}_{cr} = \operatorname{SynEncoder}\left(\left[{c}, \mathcal{T}(c)\right]\right)
\end{aligned}
\end{equation}
} where $[\cdot,\cdot]$ means the splicing operation, $X_{cr} \in \mathbb{R}^{o \times d}$ is the embeddings of the event (i.e., claim with its community response), $o$ is the maximum sequence length of PLMs. Based on the obtained response ranking, the contextually coherent posts could be potentially retained from the perspectives of the temporal and spatial relations, respectively, under the input length restriction of PLMs~\cite{devlin2019bert}.

\textbf{SemEncoder Layer.} At the semantic level, we initialize a trainable semantic encoder with the $({k+1})^{th}$ layer to the top layer of the PLMs. Then we concatenate and refine the output states of the template and the event, on top of the frozen SynEncoder, to further model the semantic interaction between the template and the event:
{
\begin{equation}\label{eq:4}
\begin{aligned}
    H = \operatorname{SemEncoder}\left(\left[X_p, X_{cr}\right]\right)
\end{aligned}
\end{equation}
} In summary, we map the simple English discrete prompt into a shared embedding space at the syntactic level by the prior knowledge of SynEncoder, which is then fed into the SemEncoder for semantic interaction with the event. 
On top of the SemEncoder, we present a prototypical verbalizer to map the output states $H^m$ of $\emph{[MASK]}$ token into the label $y$ without manual rumor-related label words for specific languages, which would be depicted in Sec.~\ref{Model Training}.

\subsection{Propagation Position Modeling}

To bridge the prompt learning and propagation structures for zero-shot rumor detection on social media, we further propose \textit{Absolute} and \textit{Relative} Propagation Position Modeling, to inject the propagation information into the tunable SemEncoder for domain-invariant structural features extraction at the semantic level. 

For Absolute Propagation Position, we exploit the propagation path of a responsive post in the propagation tree, which is complementary to its sequential counterpart~\cite{devlin2019bert}. Specifically, given a token $q$ from a post $x_i$, we treat the claim $c$ of the event as the root and use the distance of the responsive path from the current post to the root as the absolute propagation position: ${abs}_{pro}(q) = \operatorname{distance}_{tree}\left(x_i, c\right)$,
where $tree$ is the propagation structure $\mathcal{T}(c)=\langle \mathcal{G}, \overrightarrow{\mathcal{E}} \rangle$. Note that in this work, we make the tokens in the same post share the propagation position of the post in the propagation tree. Thus we update the input representation of the token $q$ for the tunable SemEncoder by summing the corresponding token embeddings in $X_{cr}$ and its absolute position embeddings, where the absolute position embeddings are trained with learnable parameters~\cite{gehring2017convolutional}.

For Relative Propagation Position, we mainly focus on the local context of a responsive post in the propagation tree as its relative propagation position. As each post in the propagation tree may trigger a set of responsive posts, we aim to capture the relative user opinions among responsive posts in such a subtree structure. Specifically, towards a post $x_i$, we consider the relative posts with five relationships in the subtree as the relative propagation position: 1) $Parent^{(+)}$; 2) $Children^{(-)}$; 3) $Siblings^{(+)}$; 4) $Siblings^{(-)}$; 5) $Itself$, where $+/-$ denotes the relative post comes earlier/later than the current post in the subtree. We then extend the self-attention computation to consider the pairwise relationships among posts in the same subtree and project the relative propagation position into the SemEncoder by drawing the practice of \citet{shaw2018self}. In this way, the relative propagation patterns in a local subtree can be captured explicitly as users share opinions towards the same subtree root, to cross-check the inaccurate information.

\subsection{Response Augmentation}
Since the model could suffer from noisy responses, we propose to enhance the prompt learning by creating additional adversarial examples. We present a new virtual response augmentation algorithm, ViRA, a variant of the virtual adversarial algorithm~\cite{miyato2018virtual}. 
To create an adversarial example, 
we apply Fast Gradient Value~\cite{rozsa2016adversarial} to approximate a worst-case perturbation, where the gradient is normalized to represent the direction that significantly decreases the model’s performance, and a norm is used to ensure the approximation is reasonable.
However, the value ranges (norms) of the embedding vectors vary among different data and models. The variance gets larger for bigger models with billions of parameters, leading to some instability of adversarial training. To this end, we first apply layer normalization~\cite{ba2016layer} on top of the frozen SynEncoder to normalize the embeddings into stochastic vectors, and then perform a mask operation to filter out the template and claim embeddings, lastly add the perturbation to the normalized embedding vectors of responsive posts. Adversarial noise enables the model to handle extensive noisy responsive posts and can be regarded as a response augmentation mechanism.

\subsection{Model Training}
\label{Model Training}
On top of the SemEncoder where the template and a event sample (i.e., a claim and its responsive posts) could be transformed into a shared semantic latent space, inspired by Prototypical Networks~\cite{snell2017prototypical, lin2021boosting}, we further introduce a prototypical verbalizer paradigm to prevent the rumor-related label words from heavily relying on the language-specific expertise words. The core idea is to utilize the representative features of instances from the same classes for encapsulating event-level semantic features instead of the language-dependent label words.

Given the $\emph{[MASK]}$ token representation $H_i^m$ of a training example $C_i$, we minimize a prototypical loss as follows:
{
\begin{equation}
    \mathcal{L}_{proto}=-log \frac{e^{ \mathcal{S}(H_i^m,l_y)}}{\sum_{y'}e^{\mathcal{S}(H_{i}^m,{l_{y'}})}}
\end{equation}} where $y$ is the ground truth of $H_i^m$, $\mathcal{S}$ denotes the normalized cosine similarity score. $l_y$ denotes the learnable prototype vectors of the class $y$, which is the cluster representative of the embedded support points belonging to the class. By optimizing the above objective function $\mathcal{L}_{proto}$, rumor features can be close to corresponding rumor prototype in semantic space and be away from the non-rumor prototype.

In addition, we adopt the contrastive loss to pull up the intra-class variance and down the inter-class variance of instances in a batch:
{
\begin{equation}
\begin{split}
    \mathcal{L}_{con}=-\frac{1}{B_{y_i}-1} \sum_{j} \mathds{1}_{[i \ne j]}\mathds{1}_{[y_i = y_j]}\\
    log\frac{e^{\mathcal{S}(H_i^m, H_j^m)}}{\sum_{j'} \mathds{1}_{[i \ne j']} e^{\mathcal{S}(H_i^m, H_{j'}^m)}}
\end{split}
\end{equation}} where $B_{y_i}$ is the number of source examples with the same label $y_i$ in the event $C_i$ in a batch, and $\mathds{1}$ is the indicator. 

We jointly train the model with the prototypical contrastive objectives: $\mathcal{L} = \alpha \mathcal{L}_{proto} + (1-\alpha) \mathcal{L}_{con}$,
where $\alpha$ is a trade-off parameter set as 0.5 in our experiments. So we generate a pseudo augmented example for $C_i$ based on response augmentation, which is again fed into the tunable SemEncoder to compute the 
new loss $\tilde{\mathcal{L}}$. Finally, we use the average loss $\mathcal{L}_{avg}= \operatorname{mean}(\mathcal{L}+\tilde{\mathcal{L}})$ for the back-propagation~\cite{collobert2011natural} with the AdamW optimizer~\cite{loshchilov2018decoupled}. We set the layer number $k$ of the SynEncoder as 6. The learning rate is initialized as 1e-5. Early stopping \cite{yao2007early} is applied to avoid overfitting.

\begin{table}[t] \Huge
\centering
\resizebox{0.475\textwidth}{!}{
\begin{tabular}{l||cc||ccc}
\hline
\multirow{2}{*}{Dataset} & \multicolumn{2}{c|}{Source} & \multicolumn{3}{c}{Target}                          \\ \cline{2-6} 
                         & TWITTER      & WEIBO        & Twitter-COVID19 & Weibo-COVID19 & CatAr-COVID19     \\ \hline \hline
\# of events             & 1154         & 4649         & 400             & 399           & 1699              \\ \hline
\# of tree nodes         & 60409        & 1956449      & 406185          & 26687         & 168276            \\ \hline
\# of non-rumors         & 579          & 2336         & 148             & 146           & 998               \\ \hline
\# of rumors             & 575          & 2313         & 252             & 253           & 701               \\ \hline
Avg. time/tree    & 389 Hours    & 1007 Hours   & 2497 Hours      & 248 Hours     & 858 Hours         \\ \hline
Avg. depth/tree          & 11.67        & 49.85        & 143.03          & 4.31          & 13.26             \\ \hline
Language                 & English      & Chinese      & English         & Chinese       & Cantonese\&Arabic \\ \hline
Domain                   & Open         & Open         & COVID-19        & COVID-19      & COVID-19          \\ \hline
\end{tabular}}
\caption{Statistics of Datasets.}
\label{tab:statistics}
\vspace{-0.5cm}
\end{table}

\section{Experiments}
\subsection{Datasets}
We utilize FOUR public datasets TWITTER, WEIBO~\cite{ma2016detecting}, Twitter-COVID19 and Weibo-COVID19~\cite{lin2022detect} for experiments. TWITTER and Twitter-COVID19 are English rumor datasets with conversation thread in tweets while WEIBO and Weibo-COVID19 are Chinese rumor datasets with the similar composition structure. Furthermore, as there are no public benchmarks available for detecting rumors in low-resource languages with propagation structure in tweets, we organized and constructed a new low-resource rumor dataset CatAr-COVID19. Specifically, we resort to two COVID-19 rumor datasets~\cite{alam2021fighting, ke2020novel}, which only contain multilingual textual claims in Cantonese and Arabic without propagation thread. We extend each claim by collecting its propagation thread via Twitter academic API in python. Finally, we annotated the claim tweets by referring to the labels of the events from the original datasets\footnote{Our code and resources will be available at \url{https://github.com/PengyaoYi/zeroRumor_AAAI}}. Statistics of the five datasets are shown in Table~\ref{tab:statistics}.

\begin{table*}[t] \Huge
\centering
\resizebox{1.\textwidth}{!}{
\begin{tabular}{l||cccccccc||cccccccc}
\hline
Source                 & \multicolumn{8}{c|}{TWITTER}                                                                                                                                                                                                         & \multicolumn{8}{c}{WEIBO}                                                                                                                                                                                                            \\ \hline
Target                 & \multicolumn{4}{c|}{Weibo-COVID19}                                                                                          & \multicolumn{4}{c|}{CatAr-COVID19}                                                                     & \multicolumn{4}{c|}{Twitter-COVID19}                                                                                        & \multicolumn{4}{c}{CatAr-COVID19}                                                                      \\ \hline
\multirow{2}{*}{Model} & \multirow{2}{*}{Acc.} & \multicolumn{1}{c|}{\multirow{2}{*}{Mac-$\emph{F}_1$}} & Rumor          & \multicolumn{1}{c|}{Non-rumor}      & \multirow{2}{*}{Acc.} & \multicolumn{1}{c|}{\multirow{2}{*}{Mac-$\emph{F}_1$}} & Rumor          & Non-rumor      & \multirow{2}{*}{Acc.} & \multicolumn{1}{c|}{\multirow{2}{*}{Mac-$\emph{F}_1$}} & Rumor          & \multicolumn{1}{c|}{Non-rumor}      & \multirow{2}{*}{Acc.} & \multicolumn{1}{c|}{\multirow{2}{*}{Mac-$\emph{F}_1$}} & Rumor          & Non-rumor      \\ \cline{4-5} \cline{8-9} \cline{12-13} \cline{16-17} 
                       &                       & \multicolumn{1}{c|}{}                        & $\emph{F}_1$             & \multicolumn{1}{c|}{$\emph{F}_1$}             &                       & \multicolumn{1}{c|}{}                        & $\emph{F}_1$             & $\emph{F}_1$             &                       & \multicolumn{1}{c|}{}                        & $\emph{F}_1$             & \multicolumn{1}{c|}{$\emph{F}_1$}             &                       & \multicolumn{1}{c|}{}                        & $\emph{F}_1$             & $\emph{F}_1$             \\ \hline \hline
Vanilla-Finetune       & 0.623                 & \multicolumn{1}{c|}{0.585}                   & 0.711          & \multicolumn{1}{c|}{0.459}          & 0.518                 & \multicolumn{1}{c|}{0.402}                   & 0.583          & 0.220          & 0.603                 & \multicolumn{1}{c|}{0.602}                   & 0.619          & \multicolumn{1}{c|}{0.585}          & 0.481                 & \multicolumn{1}{c|}{0.481}                   & 0.479          & 0.474          \\
Translate-Finetune     & 0.639                 & \multicolumn{1}{c|}{0.567}                   & 0.745          & \multicolumn{1}{c|}{0.388}          & 0.523                 & \multicolumn{1}{c|}{0.457}                   & 0.637          & 0.277          & 0.634                 & \multicolumn{1}{c|}{0.574}                   & 0.653          & \multicolumn{1}{c|}{0.495}          & 0.505                 & \multicolumn{1}{c|}{0.512}                   & 0.528          & 0.496          \\
Contrast-Finetune      & 0.656                 & \multicolumn{1}{c|}{0.582}                   & 0.759          & \multicolumn{1}{c|}{0.405}          & 0.584                 & \multicolumn{1}{c|}{0.458}                   & 0.720          & 0.196          & 0.653                 & \multicolumn{1}{c|}{0.644}                   & 0.699          & \multicolumn{1}{c|}{0.590}          & 0.562                 & \multicolumn{1}{c|}{0.561}                   & 0.571          & 0.551          \\ \hline
Adapter                & 0.644                 & \multicolumn{1}{c|}{0.600}                   & 0.737          & \multicolumn{1}{c|}{0.463}          & 0.558                 & \multicolumn{1}{c|}{0.438}                   & 0.665          & 0.211          & 0.652                 & \multicolumn{1}{c|}{0.612}                   & 0.736          & \multicolumn{1}{c|}{0.487}          & 0.548                 & \multicolumn{1}{c|}{0.556}                   & 0.605          & 0.508          \\
Parallel-Adapter       & 0.651                 & \multicolumn{1}{c|}{0.598}                   & 0.730          & \multicolumn{1}{c|}{0.467}          & 0.567                 & \multicolumn{1}{c|}{0.450}                   & 0.701          & 0.198          & 0.667                 & \multicolumn{1}{c|}{0.653}                   & 0.731          & \multicolumn{1}{c|}{0.574}          & 0.579                 & \multicolumn{1}{c|}{0.585}                   & 0.636          & 0.534          \\ \hline
Source-Prompt          & 0.664                 & \multicolumn{1}{c|}{0.648}                   & 0.722          & \multicolumn{1}{c|}{0.574}          & 0.589                 & \multicolumn{1}{c|}{0.564}                   & 0.460          & 0.669          & 0.670                 & \multicolumn{1}{c|}{0.616}                   & 0.760          & \multicolumn{1}{c|}{0.472}          & 0.599                 & \multicolumn{1}{c|}{0.565}                   & 0.688          & 0.441          \\
Translate-Prompt       & 0.650                 & \multicolumn{1}{c|}{0.489}                   & 0.776          & \multicolumn{1}{c|}{0.201}          & 0.573                 & \multicolumn{1}{c|}{0.568}                   & 0.519          & 0.617          & 0.674                 & \multicolumn{1}{c|}{0.651}                   & 0.740          & \multicolumn{1}{c|}{0.562}          & 0.604                 & \multicolumn{1}{c|}{0.542}                   & 0.374          & 0.711          \\
Soft-Prompt            & 0.652                 & \multicolumn{1}{c|}{0.574}                   & 0.756          & \multicolumn{1}{c|}{0.392}          & 0.590                 & \multicolumn{1}{c|}{0.565}                   & 0.446          & 0.683          & 0.685                 & \multicolumn{1}{c|}{0.652}                   & 0.758          & \multicolumn{1}{c|}{0.546}          & 0.609                 & \multicolumn{1}{c|}{0.575}                   & 0.518          & 0.633          \\ \hline
RPL-Cho                & 0.713                 & \multicolumn{1}{c|}{0.675}                   & 0.786          & \multicolumn{1}{c|}{0.563}          & 0.613                 & \multicolumn{1}{c|}{0.581}                   & 0.455          & 0.707 & 0.715                 & \multicolumn{1}{c|}{0.689}                   & 0.778          & \multicolumn{1}{c|}{0.601}          & 0.634                 & \multicolumn{1}{c|}{0.633}                   & \textbf{0.616} & 0.650          \\
RPL-Inv                & 0.728                 & \multicolumn{1}{c|}{0.666}                   & \textbf{0.810} & \multicolumn{1}{c|}{0.521}          & 0.601                 & \multicolumn{1}{c|}{0.592}                   & 0.473          & \textbf{0.711}          & \textbf{0.733}        & \multicolumn{1}{c|}{0.710}                   & 0.788          & \multicolumn{1}{c|}{0.632}          & 0.647                 & \multicolumn{1}{c|}{0.640}                   & 0.586          & 0.693          \\
RPL-Dep                & 0.732                 & \multicolumn{1}{c|}{0.689}                   & 0.805          & \multicolumn{1}{c|}{0.574}          & \textbf{0.640}                 & \multicolumn{1}{c|}{\textbf{0.619}}                   & 0.530          & 0.708          & 0.723                 & \multicolumn{1}{c|}{\textbf{0.711}}          & 0.771          & \multicolumn{1}{c|}{\textbf{0.650}} & 0.657                 & \multicolumn{1}{c|}{0.636}                   & 0.547          & \textbf{0.724} \\
RPL-Bre                & \textbf{0.745}        & \multicolumn{1}{c|}{\textbf{0.719}}          & 0.804          & \multicolumn{1}{c|}{\textbf{0.634}} & 0.631        & \multicolumn{1}{c|}{0.617}          & \textbf{0.544} & 0.689          & 0.727                 & \multicolumn{1}{c|}{0.697}                   & \textbf{0.793} & \multicolumn{1}{c|}{0.601}          & \textbf{0.672}        & \multicolumn{1}{c|}{\textbf{0.664}}          & 0.614          & 0.714          \\ \hline
\end{tabular}}
\caption{Rumor detection results on the target test datasets.}
\label{tab:main_results}
\vspace{-0.5cm}
\end{table*}

\subsection{Experimental Setup}
We compare our model with several state-of-the-art zero-shot rumor detection systems: 
1) \textbf{Vanilla-Finetune}: Fine-tune the model for classification by adding a task-specific linear layer with the $\emph{[CLS]}$ token on top of PLMs~\cite{devlin2019bert}; 2) \textbf{Translate-Finetune}: Utilize rumor data in source language for training and translate the claim into target languages for testing~\cite{du2021cross}; 3) \textbf{Contrast-Finetune}: We employ and extend an existing few-shot learning technique, supervised contrastive learning~\cite{lin2022detect}, to fine-tuning on the source data in the zero-shot scenario; 4) \textbf{Adapter}: Fix the parameters of PLMs and add only a few trainable parameters per task within a residual adapter~\cite{houlsby2019parameter}; 5) \textbf{Parallel-Adapter}: An adapter-based variant~\cite{he2021towards} by transferring the parallel insertion of prefix tuning into adapters; 6) \textbf{Source-Prompt}: A prompt-based tuning method~\cite{lin2021few} both trains and tests the model by prompt in source languages; 7) \textbf{Translate-Prompt}: Train on prompts in the source language and test on the target-lingual prompts after translation~\cite{zhao2021discrete}; 8) \textbf{Soft-Prompt}: Instead of discrete tokens, tunable tokens~\cite{lester2021power} are utilized as the prompt;
9) \textbf{RPL-*}: Our proposed response-aware prompt learning framework with the diverse propagation threads, i.e., chronological (Cho) and inverted (Inv) order in time sequence, depth-first (Dep) and breadth-first (Bre) order in tree structure.

In this work, we consider the most challenging case, i.e., detecting events (i.e., target) from a new domain and language. Specifically, we use the well-resourced TWITTER~\cite{ma2017detect} and WEIBO~\cite{ma2016detecting}) datasets as the source data, and Weibo-COVID19, Twitter-COVID19 and CatAr-COVID19 datasets as the target. We use accuracy and macro-averaged F1, as well as class-specific F1 scores as the evaluation metrics. 

\subsection{Rumor Detection Performance}
Table \ref{tab:main_results} shows the performance of our proposed method versus all the compared methods on the Weibo-COVID19, Twitter-COVID19 and CatAr-COVID19 datasets with pre-determined training datasets. From Table~\ref{tab:main_results}, it is observed that the performance of the baselines in the first group are obviously poor due to heavy reliance on downstream classification objectives with a task-related linear layer added on top of PLMs, which is randomly initialized and too easily overfit the source data to generalize to the target. 

The prompt-based baselines in the third group are relatively better than the adapter-based baselines in the second group though Soft-Prompt is somewhat related to the adapter style in the form of parameter tuning~\cite{he2021towards}. However, their performance are still limited to the following reasons: 1) Source-Prompt lacks cross-lingual transferability. Generally, the multilingual PLMs cannot deal well with the cross-lingual combination between the template in the source language and the claim post in the target language, where such data format is rarely seen in the pre-training stage. 2) Translate-Prompt easily suffers from error propagation of the machine translation quality, and the language-agnostic knowledge is not decoupled and transferred from the source template to the target. 3) Soft-Prompt requires abundant target rumor data for sufficient optimization, which cannot be satisfied with the zero-shot setting.

In contrast, our proposed RPL-based approaches achieve superior performance among all the baselines, which suggests their strong generalization for zero-shot transfer between different languages and different domains. It's observed that the performance of RPL-Inv is relatively better than that of RPL-Cho. We speculate that the reason is that questioning posts at the later stage of propagation could indicate a higher tendency that the claim is rumorous or not. Although achieving promising performance, RPL-Dep does not achieve the expected best performance because with the propagation of the claim there is more semantic and structural information but the noisy information is increased simultaneously, especially in the vein of relatively deep conversation or argument. Overall, RPL-Bre obtains stable and excellent performance generally among the four RPL-based variants by making full use of the subtree-structure property via breadth-first ranking and propagation position modeling for response fusion, which verifies that inaccurate information on social media can be “self-checked" by making a comparison with responsive posts towards the same topic.

\subsection{Ablation Study}
We perform ablation studies by discarding some important components of our best-performed approach RPL-Bre on CatAr-COVID19, which include 1) \textit{w/o RR}: We simply encode the claim without the Response Ranking (RR) strategies that consider the social contexts in community response. 2) \textit{w/o APP}: We discard the Absolute Propagation Position. 3) \textit{w/o RPP}: We discard the Relative Propagation Position (RPP). 4) \textit{w/o ViRA}: We neglect the Virtual Response Augmentation (ViRA) mechanism. 5) \textit{w/o HPE}: Instead of our proposed Hierarchical Prompt Encoding (HPE) mechanism, we devise our backbone as two tiers of transformers: one for encoding all the responsive posts independently, and another for processing the sequence of posts using representations from the first transformer (i.e., PLMs), where the second-tier transformer has a similar architecture to PLMs, but has only 2 layers and its parameters are initialized randomly. 6) \textit{w/o PV}: We design a manual verbalizer for label mapping, to replace the Prototypical Verbalizer (PV) for model training.

As demonstrated in Table \ref{ablation}, the ablative models suffer different degrees of such performance degradation, indicating the effectiveness of our proposed components for adapting features learned from source rumor data to that of the target. Specifically, \textit{RPL-Bre}'s performance significantly decreases without response ranking due to the lack of collective wisdom on social media. Both \textit{w/o APP} and \textit{w/o RPP} also achieve worse performance than \textit{RPL-Bre}, suggesting that both perspectives of propagation position modeling are comparably helpful to the domain-variant propagation patterns extraction in zero-shot regimes;  \textit{RPL-Bre} makes improvements over \textit{w/o ViRA},  which implies the promoting role of ViRA that enables our approach hardly compromised when the input length is limited and there may be noise in response. Moreover, \textit{w/o HPE} leads to much performance degradation, which implies the prompt encoding framework ingeniously reserves the prior syntactic and semantic knowledge from the PLMs and contributes more accurate zero-shot rumor predictions with language disentanglement. Compared with \textit{RPL-Bre}, the performance of \textit{w/o PV} also significantly decreases, highlighting the importance and complementary of the prototypical paradigm in our framework for language and domain adaptation. 


{
\begin{table}[t] \Huge
\centering
\resizebox{0.35\textwidth}{!}{
\begin{tabular}{l||ll||ll}
\hline
Source           & \multicolumn{2}{c|}{TWITTER}                           & \multicolumn{2}{c}{WEIBO}                             \\ \hline
Model            & \multicolumn{1}{c}{Acc.} & \multicolumn{1}{c|}{Mac-$\emph{F}_1$} & \multicolumn{1}{c}{Acc.} & \multicolumn{1}{c}{Mac-$\emph{F}_1$} \\ \hline \hline
RPL-Bre          & 0.631                    & 0.617                       & 0.672                    & 0.664                      \\ \hline
RPL-Bre w/o RR   & 0.605                    & 0.598                       & 0.613                    & 0.611                      \\
RPL-Bre w/o APP  & 0.622                    & 0.607                       & 0.626                    & 0.624                      \\
RPL-Bre w/o RPP  & 0.610                    & 0.601                       & 0.633                    & 0.632                      \\
RPL-Bre w/o ViRA & 0.626                    & 0.612                       & 0.644                    & 0.634                      \\
RPL-Bre w/o HPE  & 0.571                    & 0.451                       & 0.581                    & 0.433                      \\
RPL-Bre w/o PV  & 0.592                    & 0.589                       & 0.621                    & 0.617                       \\ \hline
\end{tabular}}
\caption{Ablation studies on our proposed model.}
\vspace{-0.5cm}
\label{ablation}
\end{table}}

\subsection{Early Detection}
Early alerts of rumors can prevent the wide-spreading of rumorous contents. By setting detection checkpoints of ``delays" that can be either the count of reply posts or the time elapsed since the first posting, only contents posted no later than the checkpoints is available for model evaluation. The performance is evaluated by Macro F1 obtained at each checkpoint. To satisfy each checkpoint, we incrementally scan test data in order of time until the target time delay or post volume is reached.

Figure~\ref{fig:early_detection} shows the early detection performance of our approach versus Soft-Prompt, PLAN, STANKER, BiGCN and RvNN at various deadlines. To make fair comparisons, the inputs of all baselines are encoded with the same multilingual PLM. We observe that our proposed RPL-based approach outperforms other baselines throughout the whole lifecycle, and reaches a relatively high Macro F1 score at a very early period after the initial broadcast. One interesting phenomenon is that our method only needs about 20 posts on CatAr-COVID19 and 4 hours on Twitter-COVID19, to achieve the saturated performance, indicating the advanced response fusion strategy and remarkably superior early detection performance of our method.

{
\setlength{\abovecaptionskip}{-0.01cm}
\setlength{\belowcaptionskip}{-0.01cm}
\begin{figure}[t]
\centering
\subfigure{
\begin{minipage}[t]{0.5\linewidth}
\centering
\scalebox{0.85}{\includegraphics[width=5cm]{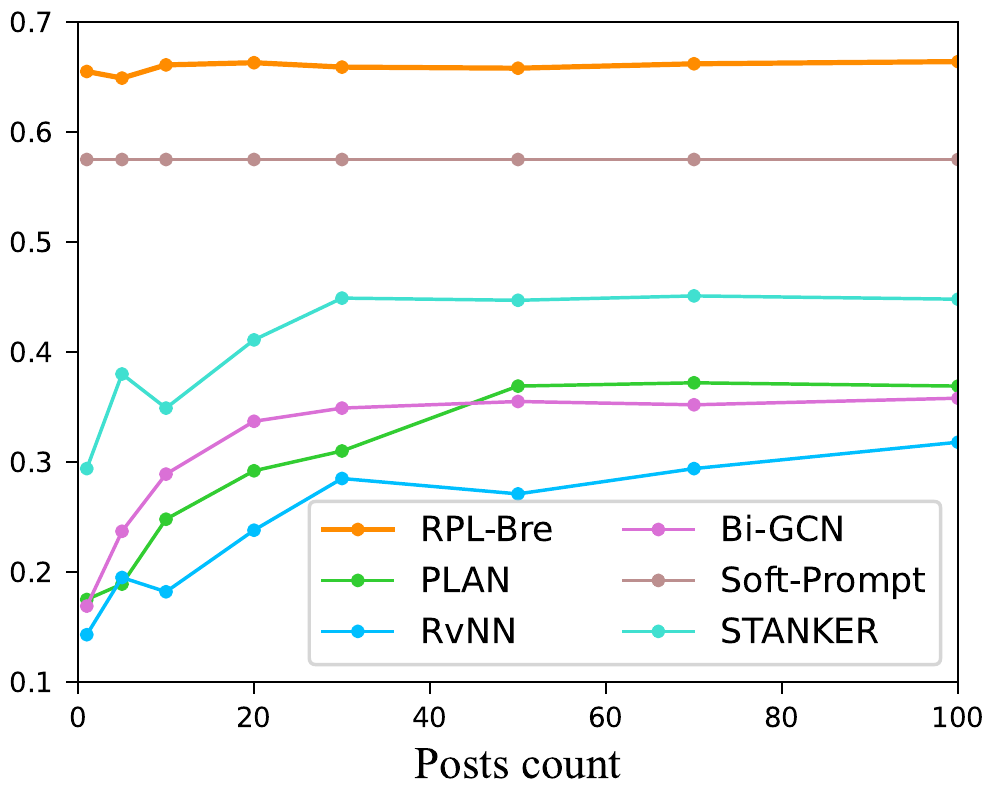}}
\end{minipage}%
}%
\subfigure{
\begin{minipage}[t]{0.5\linewidth}
\centering
\scalebox{0.85}{\includegraphics[width=5cm]{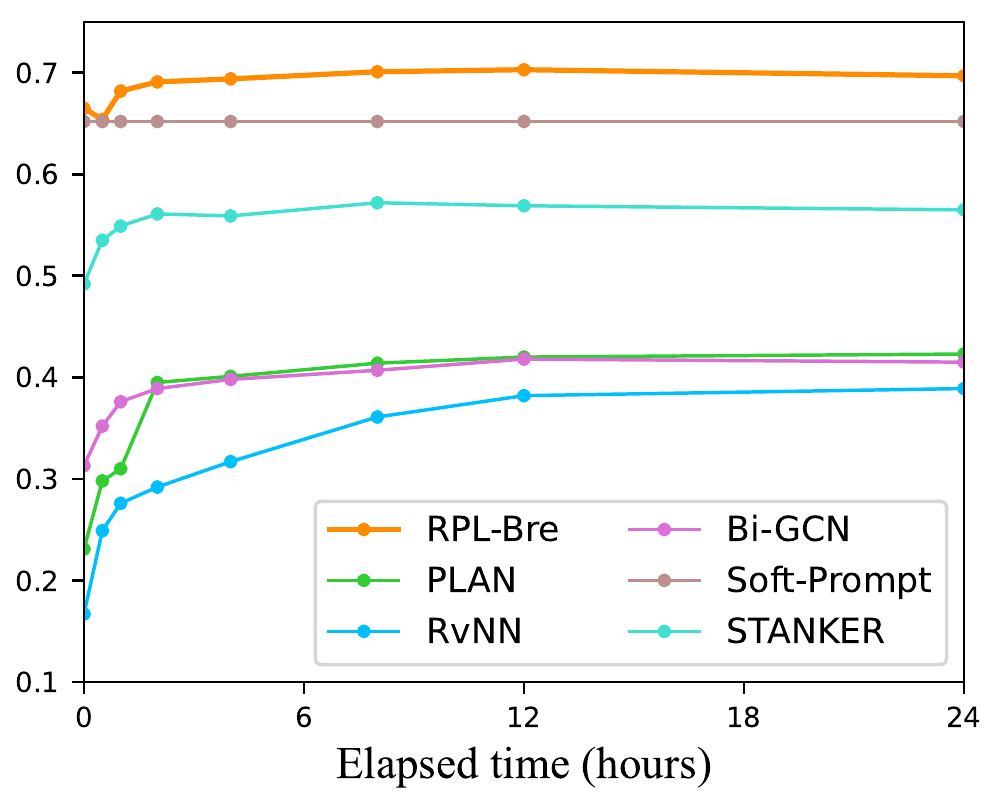}}
\end{minipage}%
}%
\centering
\caption{Early detection performance at different checkpoints of posts count (or elapsed time) on CatAr-COVID19 (left) and Twitter-COVID19 (right) datasets.}
\label{fig:early_detection}
\vspace{-0.1cm}
\end{figure}}

{
\setlength{\abovecaptionskip}{-0.01cm}
\setlength{\belowcaptionskip}{-0.01cm}
\begin{figure}[t]
\centering
\subfigure{
\begin{minipage}[t]{0.5\linewidth}
\centering
\scalebox{0.83}{\includegraphics[width=5cm]{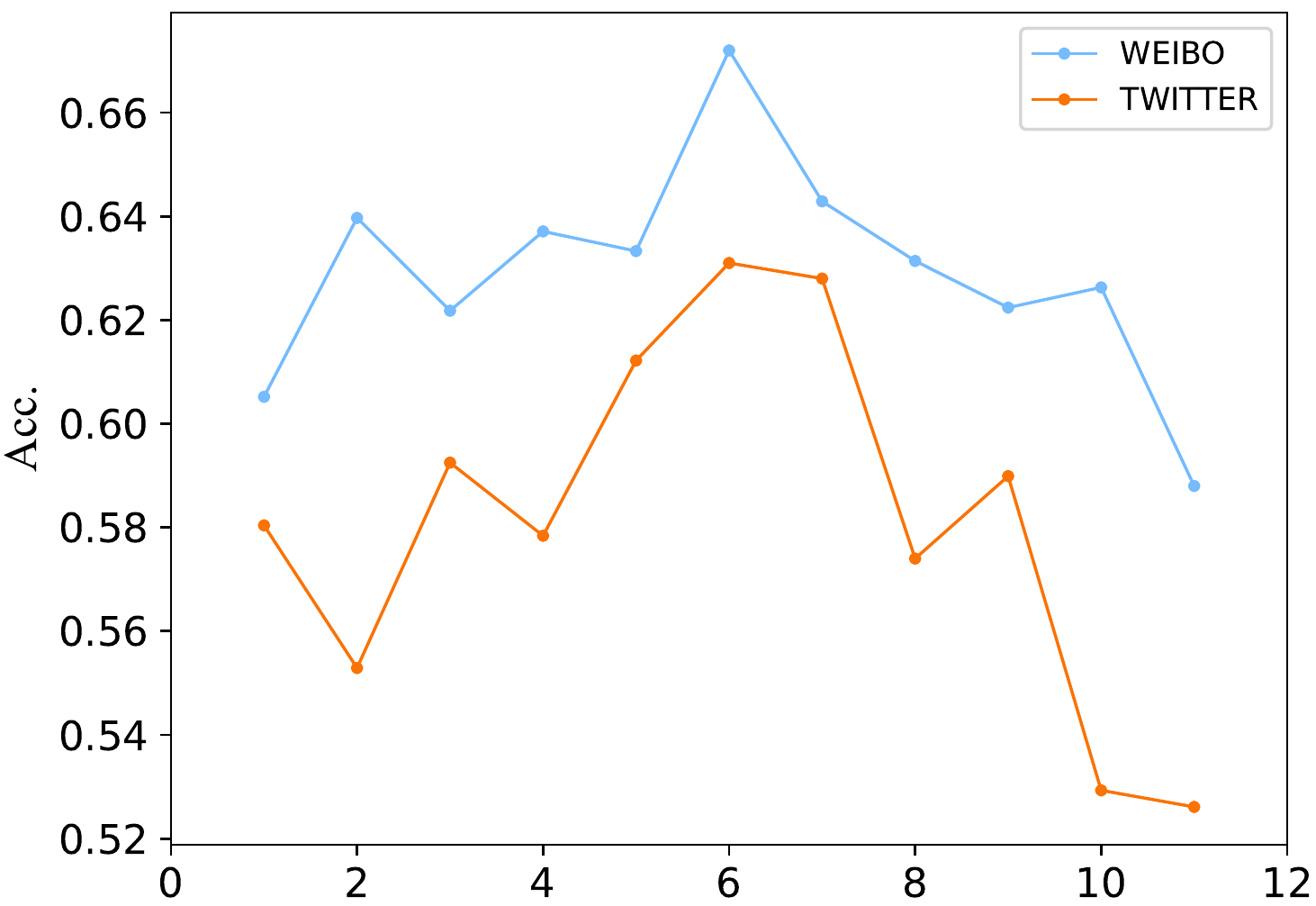}}
\end{minipage}%
}%
\subfigure{
\begin{minipage}[t]{0.5\linewidth}
\centering
\scalebox{0.82}{\includegraphics[width=5cm]{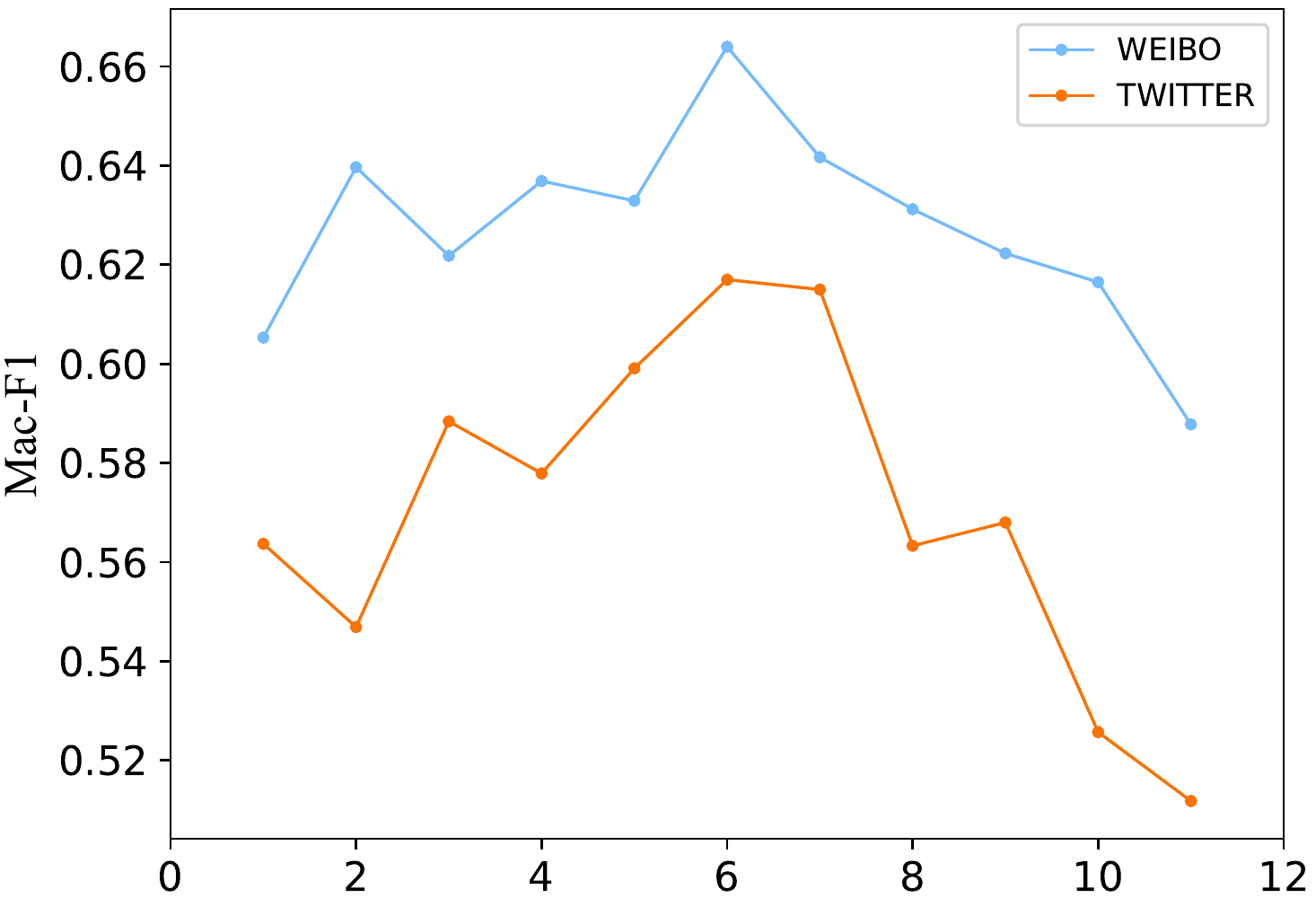}}
\end{minipage}%
}%
\centering
\caption{Effect of the layer number $k$ of SynEncoder with Accuracy (left) and Macro F1 (right).}
\label{fig:layer_split}
\vspace{-0.5cm}
\end{figure}}

\subsection{Discussion}
Figure~\ref{fig:layer_split} shows the effect of layer number of the SynEncoder on zero-shot rumor detection performance, with the CatAr-COVID19 as the target, TWITTER and WEIBO as the source data, respectively. We can observe that when the SynEncoder is initialized with the lower 4 layers of PLMs, it is still biased to specific languages due to the surface features mainly learned. Since PLMs could unearth rich linguistic features at the lower 6 layers, the best performance is obtained when $k$ is set to 6 (i.e., the setting in our model), which is in line with the finding of \citet{jawahar2019does}. After that, as $k$ continues to increase, although the capacity to decouple shared semanteme from specific linguistic features is enhanced, the number of SemEncoder layers with prior semantic knowledge activated for the interaction of prompts and events decreases, thus the generalization ability of the model to rumor data in different domains is limited, resulting in a fluctuated decline of performance.

\section{Conclusion and Future Work}
In this paper, we propose a zero-shot Response-aware Prompt Learning framework to bridge language and domain gaps in rumor detection. We present a prompt-based approach to avoid the reliance on language-specific rumor prompt engineering, with effective response fusion strategies to incorporate influential and structural propagation threads for domain adaptation. Results on three real-world benchmarks confirm the advantages of our zero-shot detection model. For future work, we plan to study specialized PLMs for rumor detection to better utilize the wisdom of crowds and circumvent the sequence length limit, then collect and apply our model to more languages and domains.

\setcounter{secnumdepth}{0}
\section{Acknowledgments}
Work partially done when Hongzhan Lin was an intern at Tencent AI Lab. This work was partially supported by Hong Kong RGC ECS (Ref. 22200722) and HKBU One-off Tier 2 Start-up Grant (Ref. RCOFSGT2/20-21/SCI/004).

\bibliography{aaai23}

\appendix
\section{Datasets}
The focus of this work, as well as in many previous studies~\cite{ma2018rumor, khoo2020interpretable, bian2020rumor, lin2021rumor}, is rumors on social media, not just the "fake news" strictly defined as a news article published by a news outlet that is verifiably false~\cite{zubiaga2018detection}. We utilize FOUR public datasets TWITTER, WEIBO~\cite{ma2016detecting}, Twitter-COVID19 and Weibo-COVID19~\cite{lin2022detect} for experiments. TWITTER and Twitter-COVID19 are English rumor datasets with conversation threads in tweets while WEIBO and Weibo-COVID19 are rumor datasets in Chinese Simplified, with the similar composition structure as TWITTER and Twitter-COVID19. Furthermore, as there are no public benchmarks available for detecting rumors in low-resource languages with a propagation tree structure in tweets, we organized and constructed a new low-resource rumor dataset CatAr-COVID19. Specifically, we resort to two COVID-19 rumor datasets~\cite{alam2021fighting, ke2020novel}, which only contain multilingual textual claims in Cantonese and Arabic without a propagation thread. We extend each claim by collecting its propagation threads via Twitter academic API with a twarc2 package\footnote{\url{https://twarc-project.readthedocs.io/en/latest/twarc2_en_us/}} in python. Finally, we annotated the claim tweets by referring to the labels of the events from the original datasets, where rumors contain facts or misinformation remaining to be verified while non-rumors do not. The detailed statistics of the CatAr-COVID19 dataset are shown in Table~\ref{statistics of CatAr}. 

\section{Implementation Details} 
Since the focus of this paper is primarily on better leveraging the prompt learning for language and domain adaptation in zero-shot regimes, we implement the $\text{XLM-R}_{\textit{Base}}$ (Layer number = 12, Hidden dimension = 768, Attention head = 12, 270M params) as the PLM backbone of our Hierarchical Prompt Encoding mechanism and all the baselines in our experiments. We set the layer number $k$ of SynEncoder as 6 for the Hierarchical Prompt Encoding. To choose the trade-off parameter $\alpha$, we conducted a grid search within the range $[0, 1]$ and picked the number $0.5$ that results in best performance on Macro F1. Parameters are updated through back-propagation~\cite{collobert2011natural} with the AdamW optimizer~\cite{loshchilov2018decoupled}. The learning rate is initialized as 1e-5. Early stopping \cite{yao2007early} is applied to avoid overfitting. We set the total batch size to 16. The max sequence length of the syntactic encoder is set to 512. We use accuracy and macro-averaged F1 score, as well as class-specific F1 score as the evaluation metrics. For each experiment reported in this work, we use 10 different random seeds to run the model and report the average results. We hold out 10\% of the target test datasets for tuning the hyper-parameters. The number of total trainable parameters is 278,197,248 for our model. The number of total parameters is 513,123,200 for our model. We run all of our experiments on one single NVIDIA Tesla V100 GPU. We implement our model with pytorch\footnote{\url{pytorch.org}} and HuggingFace Transformers~\cite{wolf2020transformers}, and the template prompt is in English generated by the code released by~\cite{gao2021making}.

\begin{table}[] \scriptsize
\centering
{\begin{tabular}{l||cc}
\hline
\multirow{2}{*}{Dataset} & \multicolumn{2}{c}{CatAr-COVID19} \\ \cline{2-3} 
                         & Cantonese       & Arabic          \\ \hline \hline
\# of events             & 1481            & 218             \\ \hline
\# of tree nodes         & 68490           & 99786           \\ \hline
\# of non-rumors         & 920             & 78              \\ \hline
\# of rumors             & 561             & 140             \\ \hline
Avg. time length/tree    & 668 Hours       & 2154 Hours      \\ \hline
Avg. depth/tree          & 9.98            & 35.54           \\ \hline
\end{tabular}}
\caption{Statistics of CatAr-COVID19 dataset.}
\label{statistics of CatAr}
\end{table}

\section{RPL Algorithm}
Algorithm~\ref{algorithm} presents the training and testing procedure of our approach.

\begin{algorithm}[t]
  \caption{\textbf{Response-aware Prompt Learning}}
  \label{algorithm}
  \begin{algorithmic}[1]
    \Require
      A set of events $C^s$ in the source domain and language; A set of events $C^t$ in the target domain and language.
    \Ensure
      Assign rumor labels $y$ to given unlabeled events $C^t$ in the target domain and language.
    \State \textbf{\textit{Train on $C^s$:}}
    \State \textbf{for} each mini-batch $B$ of the source events $C^s$ \textbf{do}:
    \State \quad Pass $C^s$ to the model and then obtain its output states $H_s^m$ of \emph{[MASK]} token on top of SemEncoder. 
    \State \quad Compute the prototypical contrastive loss $\mathcal{L}$ for source data.
    \State \quad Response augmentation for the training data and perform the second forward for the new loss $\tilde{\mathcal{L}}$.
    \State \quad Jointly optimize all trainable parameters of the model using the average loss $\mathcal{L}_{avg}= \operatorname{mean}(\mathcal{L}+\tilde{\mathcal{L}})$.
    \State \textbf{end for}
    \State \textbf{\textit{Test on $C^t$:}}
    \State Pass $C^t$ to the model and then obtain its output states $H_t$ on top of SemEncoder.
    \State The similarity score between the output states $H_t^m$ of \emph{[MASK]} and prototypes would be used to predict labels.
  \end{algorithmic}
\end{algorithm}

\section{Supplemental Experiments}
\subsection{Zero-shot Rumor Detection on CatAr-COVID19}
We pick the Arabic and Cantonese claims, respectively, with their propagation thread in the CatAr-COVID19 dataset to conduct a supplementary experiment for our proposed model, using the WEIBO dataset as the source data, as shown in Table~\ref{ARABIC}. Different from Cantonese or Chinese Traditional which have many cognates with Chinese Simplified, generally, the complicated grammar and the "long and obscure" sentences are the characteristics of Arabic, as shown in Figure~\ref{fig:wordcloud}. We could find that our models perform well on the low-resource data in Arabic corresponding to COVID-19, leveraging the source data in Chinese Simplified. Due to the limited resource for the target language, the volume of Arabic data is relatively smaller. Therefore, we plan to collect more related data for the low-resource languages for providing comprehensive guidance for future rumor detection about breaking events on social media. Furthermore, we could observe that our models trained on Chinese Simplified data obtain comparable performance on Cantonese, which is mainly written in Chinese Traditional, not only popular in Guangdong Province, Hong Kong, Macau even Taiwan, but also widely used by overseas Chinese people. Though Cantonese is relatively closer to Chinese, it is joined with British English and combined with the local culture, as the examples shown in Table~\ref{example}. How to use the propagation structure to further improve the performance on breaking events in this language needs more systematic and targeted research.

\subsection{Zero-shot Cross-lingual Rumor Detection}
In this section, we evaluate our proposed framework with different source datasets to discuss the zero-shot settings in our experiments. Considering the cross-domain and cross-lingual settings concurrently in the main experiments, we also conduct an experiment in only cross-lingual settings. Specifically, for the TWITTER as the target data, we utilize the WEIBO dataset as the source data with rich annotation. In terms of WEIBO as the target data, we set the WEIBO dataset as the source data. Table~\ref{cross-lingual} depicted the results in zero-shot cross-lingual settings. It can be seen from the results that the RPL-Bre performs generally better in cross-lingual settings than the other variants of our model, which reaffirms that the breadth-first ranking is the stable choice in terms of cross-lingual scenarios across different datasets.

{
\begin{figure}[t]
\centering
\subfigure[Rumor]{
\begin{minipage}[t]{0.5\linewidth}
\centering
\scalebox{0.75}{\includegraphics[width=5cm]{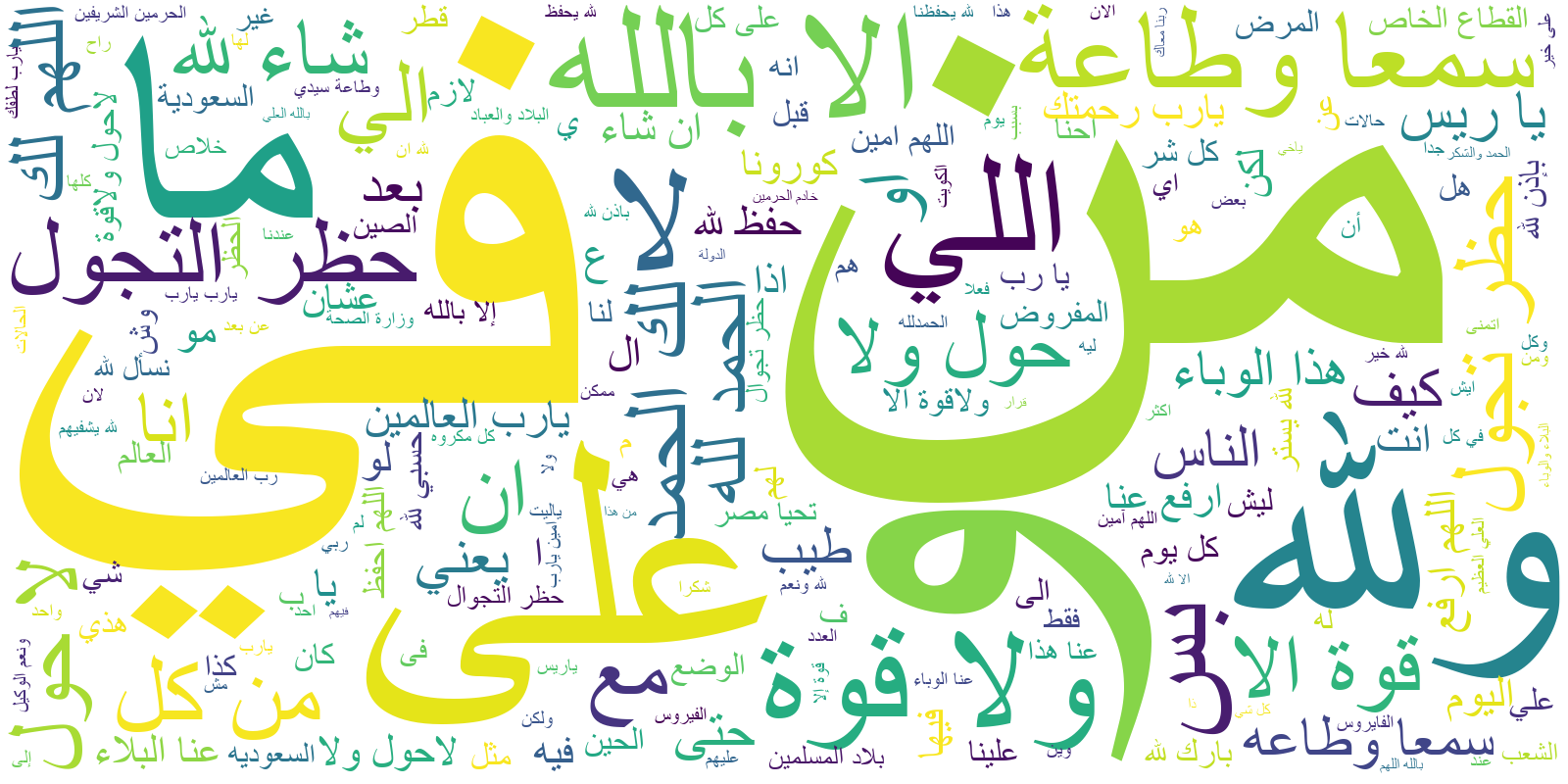}}
\end{minipage}%
}%
\subfigure[Non-rumor]{
\begin{minipage}[t]{0.5\linewidth}
\centering
\scalebox{0.75}{\includegraphics[width=5cm]{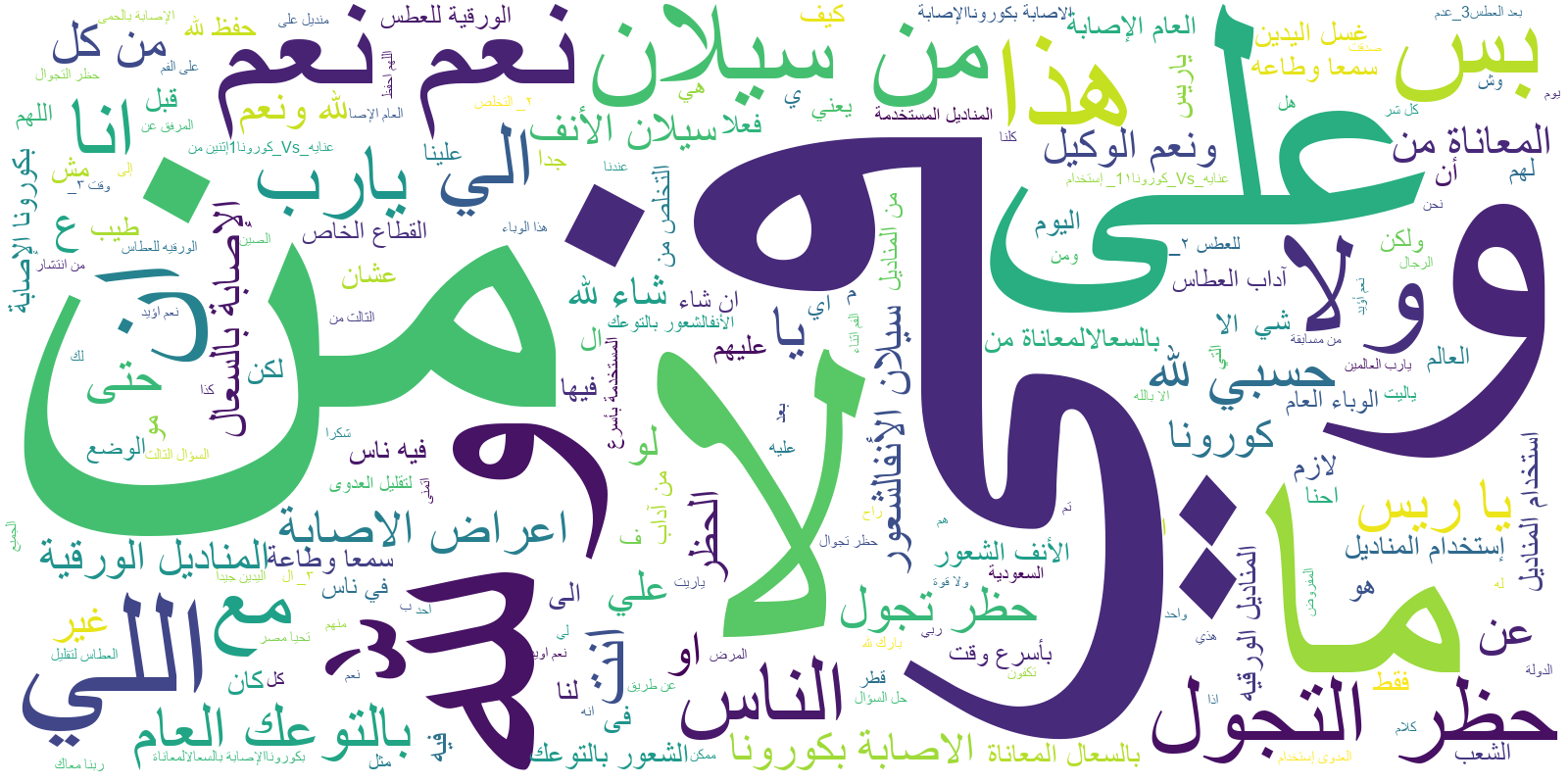}}
\end{minipage}%
}%
\centering
\caption{Word clouds of rumor and non-rumor data generated from Arabic COVID-19 data.}
\label{fig:wordcloud}
\end{figure}}

\begin{table}[t] \Huge
\centering
\resizebox{0.475\textwidth}{!}{
\begin{tabular}{l||cccccccc}
\hline
Source                 & \multicolumn{8}{c}{Chinese Simplified}                                                                                                                                                                   \\ \hline
Target                 & \multicolumn{4}{c|}{Arabic}                                                                                   & \multicolumn{4}{c}{Cantonese}                                                            \\ \hline
\multirow{2}{*}{Model} & \multirow{2}{*}{Acc.} & \multicolumn{1}{c|}{\multirow{2}{*}{Mac-$\emph{F}_1$}} & Rumor & \multicolumn{1}{c||}{Non-rumor} & \multirow{2}{*}{Acc.} & \multicolumn{1}{c|}{\multirow{2}{*}{Mac-$\emph{F}_1$}} & Rumor & Non-rumor \\ \cline{4-5} \cline{8-9} 
                      &                       & \multicolumn{1}{c|}{}                        & $\emph{F}_1$    & \multicolumn{1}{c||}{$\emph{F}_1$}        &                       & \multicolumn{1}{c|}{}                        & $\emph{F}_1$    & $\emph{F}_1$        \\ \hline
RPL-Cho                & 0.796                 & \multicolumn{1}{c|}{0.760}                   & 0.853 & \multicolumn{1}{c||}{0.667}     & 0.683                 & \multicolumn{1}{c|}{0.674}                   & 0.619 & 0.729     \\
RPL-Inv                & 0.788                 & \multicolumn{1}{c|}{0.757}                   & 0.845 & \multicolumn{1}{c||}{0.669}     & 0.678                 & \multicolumn{1}{c|}{0.676}                   & 0.647 & 0.704     \\
RPL-Dep                & 0.798                 & \multicolumn{1}{c|}{0.756}                   & 0.854 & \multicolumn{1}{c||}{0.658}     & 0.691                 & \multicolumn{1}{c|}{0.676}                   & 0.606 & 0.746     \\
RPL-Bre                & 0.791                 & \multicolumn{1}{c|}{0.752}                   & 0.853 & \multicolumn{1}{c||}{0.651}     & 0.689                 & \multicolumn{1}{c|}{0.679}                   & 0.624 & 0.734     \\ \hline
\end{tabular}}
\caption{Results of Zero-shot Rumor Detection on Arabic and Cantonese COVID-19 data, respectively.}
\label{ARABIC}
\end{table}

\begin{table*}[t]
    \centering
    \resizebox{0.98\textwidth}{!}{\begin{tabular}{p{1.0\linewidth}}
     \toprule 
     $[$\textbf{Label:} RUMOR$]$ \textbf{Claim:} \textit{\begin{CJK*}{UTF8}{bsmi} 蘋果【武漢肺炎】英斥\$1.5億買中國製檢測工具竟無效 廠商辯稱：英方誤解效用//英國斥資2000萬美元向兩間中國公司購買了共200萬套武漢肺炎抗體檢測工具，但後被揭發檢測工具無效不過中國廠商辯稱是英方誤解或誇大了檢測工具的…\end{CJK*}}\\
     \textbf{Response1:} \textit{\begin{CJK*}{UTF8}{bsmi} 明明係你唔識用，又老屈中國？不知所謂\end{CJK*}} 
    \\
    \textbf{Response2:} \textit{\begin{CJK*}{UTF8}{bsmi} 梗係唔識用啦，因為連大陸自己都唔識用��自湖北解封後，大陸多個省市已再有湖北傳入確診個案出現\end{CJK*}} \\
    (...)\\
    \midrule \midrule 
     $[$\textbf{Label:} NON-RUMOR$]$ \textbf{Claim:} \textit{\begin{CJK*}{UTF8}{bsmi}根據世衛邏輯，用食物產地做標籤是在歧視當地，所以必需更改食品名稱免歧視我要一份新型牛肉，兩隻新型龍蝦\end{CJK*}} \\
      \textbf{Response1:} \textit{\begin{CJK*}{UTF8}{bsmi} 去到餐廳咁樣落單應該會被人藐爆\end{CJK*}} 
    \\
    \textbf{Response2:} \textit{\begin{CJK*}{UTF8}{bsmi} 搞到咁，無晒食慾\end{CJK*}} \\
    (...)\\
    \bottomrule  
     \end{tabular}}
     \caption{Cantonese Examples from CatAr-COVID19. } 
     \label{example}
     \vspace{-0.15cm}
\end{table*}

\begin{table}[t] \Huge
\centering
\resizebox{0.475\textwidth}{!}{
\begin{tabular}{l||cccc||cccc}
\hline
Source                 & \multicolumn{4}{c|}{WEIBO}                                                                             & \multicolumn{4}{c}{TWITTER}                                                                            \\ \hline
Target                 & \multicolumn{4}{c|}{TWITTER}                                                                           & \multicolumn{4}{c}{WEIBO}                                                                              \\ \hline
\multirow{2}{*}{Model} & \multirow{2}{*}{Acc.} & \multicolumn{1}{c|}{\multirow{2}{*}{Mac-$\emph{F}_1$}} & Rumor          & Non-rumor      & \multirow{2}{*}{Acc.} & \multicolumn{1}{c|}{\multirow{2}{*}{Mac-$\emph{F}_1$}} & Rumor          & Non-rumor      \\ \cline{4-5} \cline{8-9} 
                      &                       & \multicolumn{1}{c|}{}                        & $\emph{F}_1$             & $\emph{F}_1$             &                       & \multicolumn{1}{c|}{}                        & $\emph{F}_1$             & $\emph{F}_1$             \\ \hline \hline
RPL-Cho                & 0.691                 & \multicolumn{1}{c|}{0.689}                   & 0.664          & 0.714          & 0.680                 & \multicolumn{1}{c|}{0.679}                   & 0.689          & 0.670          \\
RPL-Inv                & 0.653                 & \multicolumn{1}{c|}{0.653}                   & 0.647          & 0.659          & {0.696}        & \multicolumn{1}{c|}{{0.694}}          & {0.720} & 0.668          \\
RPL-Dep                & 0.688                 & \multicolumn{1}{c|}{0.685}                   & 0.655          & {0.716}          & 0.656                 & \multicolumn{1}{c|}{0.656}                   & 0.659          & 0.653          \\
RPL-Bre                & {0.696}        & \multicolumn{1}{c|}{{0.696}}          & {0.692} & 0.699 & 0.684                 & \multicolumn{1}{c|}{0.684}                   & 0.681          & {0.687} \\ \hline
\end{tabular}}
\caption{Results of Zero-shot Cross-lingual Rumor Detection.}
\label{cross-lingual}
\end{table}

\begin{table}[t] \Huge
\centering
\resizebox{0.475\textwidth}{!}{
\begin{tabular}{l||cccc||cccc}
\hline
Source                 & \multicolumn{4}{c|}{WEIBO}                                                                             & \multicolumn{4}{c}{TWITTER}                                                                            \\ \hline
Target                 & \multicolumn{4}{c|}{Weibo-COVID19}                                                                     & \multicolumn{4}{c}{Twitter-COVID19}                                                                    \\ \hline
\multirow{2}{*}{Model} & \multirow{2}{*}{Acc.} & \multicolumn{1}{c|}{\multirow{2}{*}{Mac-$\emph{F}_1$}} & Rumor          & Non-rumor      & \multirow{2}{*}{Acc.} & \multicolumn{1}{c|}{\multirow{2}{*}{Mac-$\emph{F}_1$}} & Rumor          & Non-rumor      \\ \cline{4-5} \cline{8-9} 
                      &                       & \multicolumn{1}{c|}{}                        & $\emph{F}_1$             & $\emph{F}_1$             &                       & \multicolumn{1}{c|}{}                        & $\emph{F}_1$             & $\emph{F}_1$             \\ \hline \hline
RPL-Cho                & 0.869                 & \multicolumn{1}{c|}{0.857}                   & 0.898          & 0.816          & 0.676                 & \multicolumn{1}{c|}{0.551}                   & 0.789          & 0.313          \\
RPL-Inv                & {0.904}        & \multicolumn{1}{c|}{{0.896}}          & {0.925} & {0.866} & 0.654                 & \multicolumn{1}{c|}{0.457}                   & 0.784          & 0.130          \\
RPL-Dep                & 0.884                 & \multicolumn{1}{c|}{0.876}                   & 0.907          & 0.845          & 0.673                 & \multicolumn{1}{c|}{{0.610}}          & 0.766          & {0.454} \\
RPL-Bre                & 0.881                 & \multicolumn{1}{c|}{0.868}                   & 0.910          & 0.825          & {0.677}        & \multicolumn{1}{c|}{0.540}                   & {0.791} & 0.290          \\ \hline
\end{tabular}}
\caption{Results of Zero-shot Cross-domain Rumor Detection.}
\label{cross-domain}
\vspace{-0.15cm}
\end{table}

\subsection{Zero-shot Cross-domain Rumor Detection}
In this section, we also conduct an experiment in only cross-domain settings to evluate our proposed framework. Specifically, for the Weibo-COVID19 as the target data, we utilize the WEIBO dataset as the source data with rich annotation. In terms of Twitter-COVID19 as the target data, we set the TWITTER dataset as the source data. With WEIBO as the source data, our model can  achieve ranging from 86.9\% Accuracy and 85.7\% Macro F1 score to 90.4\% Accuracy and 89.6\% Macro F1 score of rumor detection performance on the target data Weibo-COVID19, which indicates that our superior capacity in zero-shot cross-domain rumor detection in Chinese language. However, the overall performance on Twitter-COVID19 is relatively worse with the TWITTER as the source dataset. We speculate the reason is that the number of the events in TWITTER dataset is smaller than that in WEIBO, in which our model could achieve about 72.3\% Accuracy and 71.1\% Macro F1 score among the variants of response ranking. It further demonstrates the key insight to bridge the low-resource gap is to relieve the limitation imposed by the specific language resource dependency besides the specific domain. Our proposed prompt learning framework could alleviate the low-resource issue of rumor detection as well as reduce the heavy reliance on datasets annotated with specific domain and language knowledge, which is enable to leverage the knowledge from WEIBO instead of just TWITTER to detect rumors in Twitter-COVID19 for better performance.

{
\begin{figure}[t]
    \centering
    \resizebox{0.35\textwidth}{!}{\includegraphics{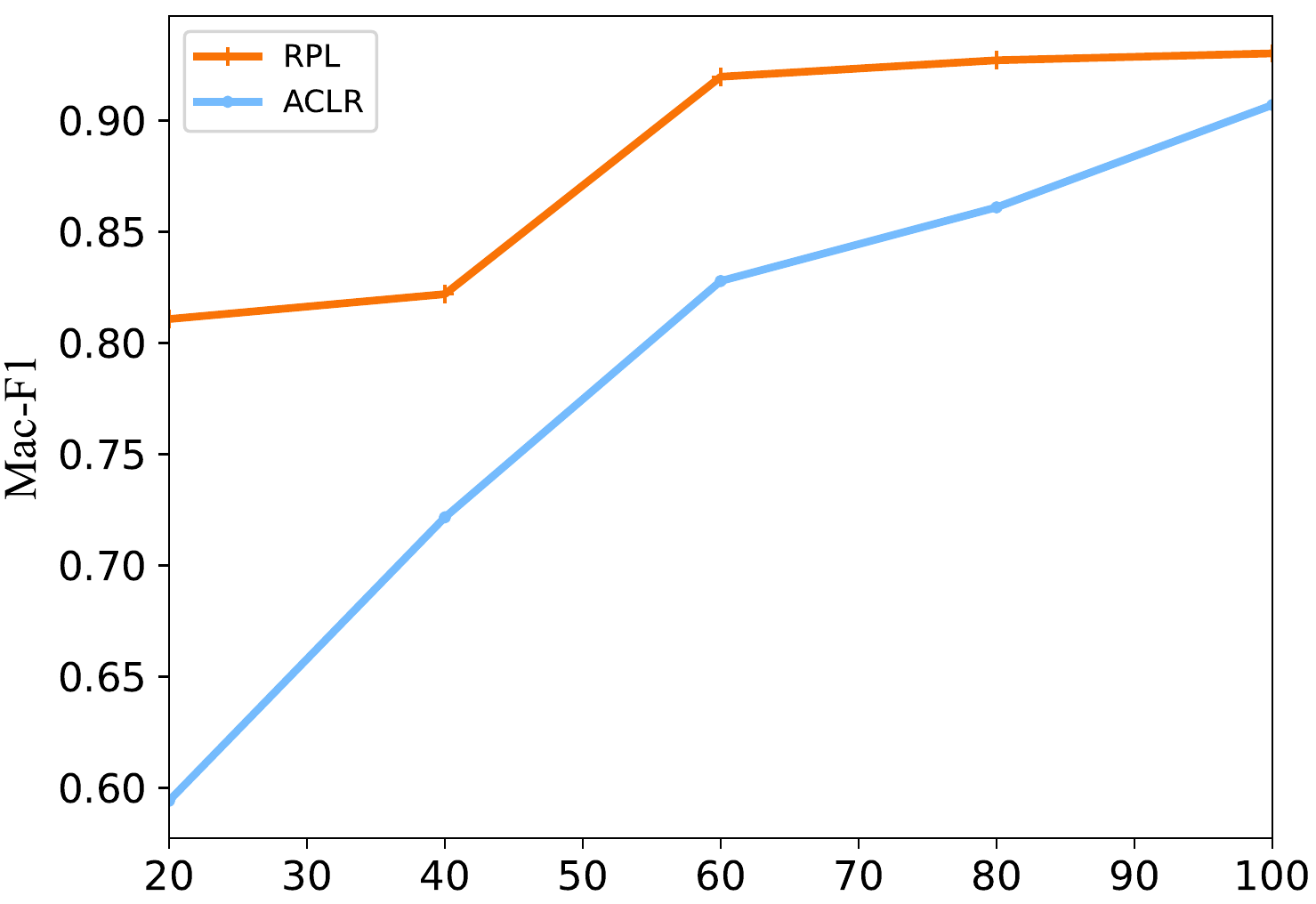}}
    \caption{Effect of target training data size on Weibo-COVID19.}
    \label{fig:training_size1}
    \vspace{-0.1cm}
\end{figure}}

{
\begin{figure}[]
    \centering
    \resizebox{0.35\textwidth}{!}{\includegraphics{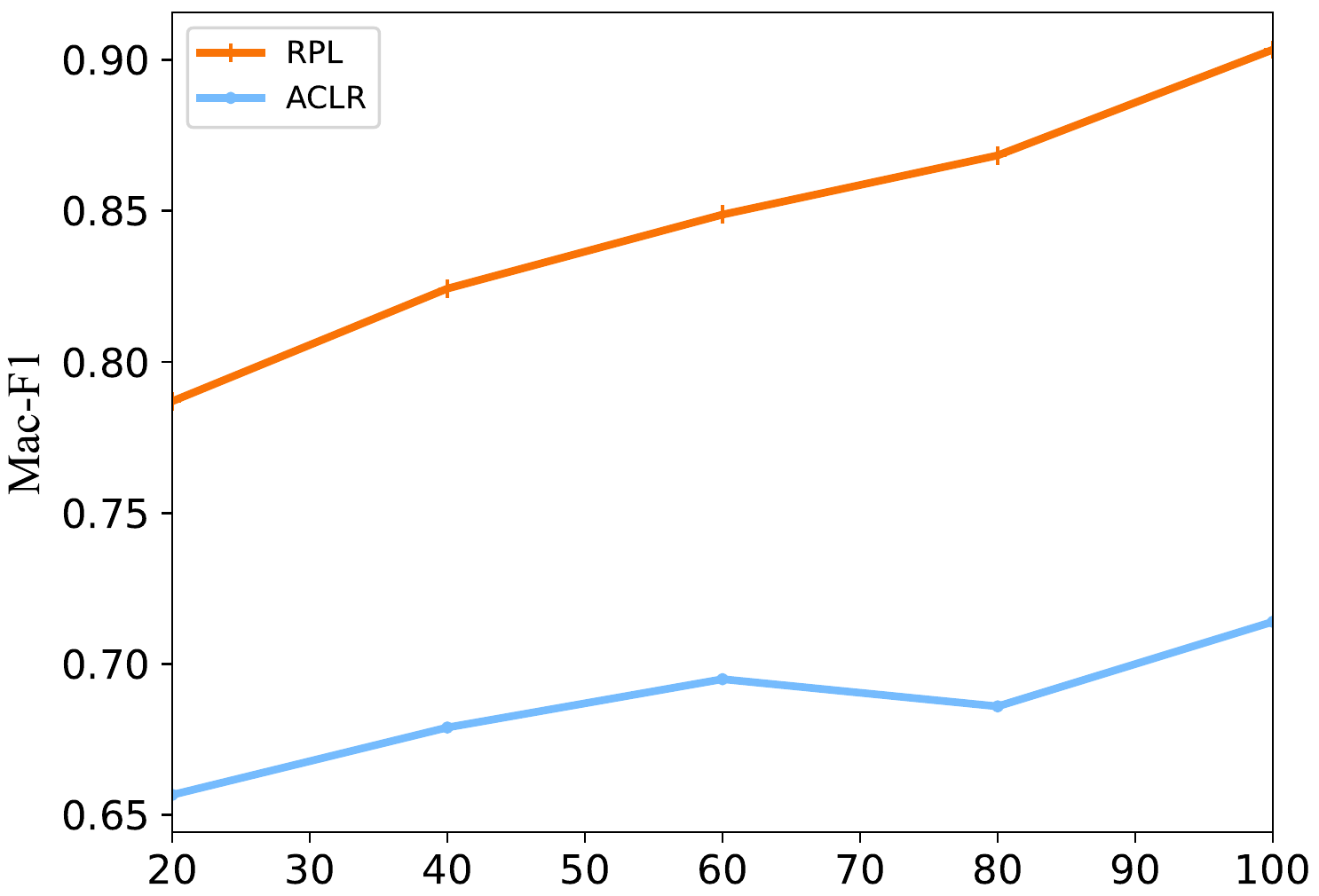}}
    \caption{Effect of target training data size on Twitter-COVID19.}
    \label{fig:training_size2}
    \vspace{-0.15cm}
\end{figure}}

\subsection{Effect of Target Training Data Size for Few-shot Rumor Detection}
To make a fair comparison with the few-shot rumor detection model ACLR proposed by \citet{lin2022detect}, we also evaluate our model versus ACLR in few-shot settings to investigate the performance of our model when the target training data size increases. Figure~\ref{fig:training_size1} and Figure~\ref{fig:training_size2} show the effect of target training data size on Weibo-COVID19 and Twitter-COVID19. We randomly choose training data with a certain proportion from target data and use the rest set for evaluation. We use the cross-domain and cross-lingual settings concurrently for model training, the same as the main experiments. Results show that with the increase of training data size, the performance gradually increases. It can be observed that even when only 20 target data are used for training, our model can still achieve more than approximately 81\% and 78\% rumor detection performance (Macro F1 score) on two target datasets Weibo-COVID19 and Twitter-COVID19 respectively, compared with the 59\% and 65\% performance in terms of ACLR, which further proves RPL has stronger applicability for improving rumor detection on social media under low-resource regimes.

\subsection{Limitations}
For more targeted work in the future, we conduct the limitations of our work based on error cases where our model can not predict the correct label of the claim:
\begin{itemize}
\item Due to the limitation of the input sequence length in PLMs, models based on PLMs could only process about 30 responsive posts for each claim according to statistics. Though we have tried the two-tier Transformer architecture to alleviate the issue, it would lead to feature loss during the transformation from token-level features at the first tier to post-level features at the second tier. Therefore, it does not obtain a more satisfactory performance than our proposed model RPL in the zero-shot scenario, which employs the response ranking to highlight the potentially evidential and contextual posts. It's necessary to study specialized PLMs for the rumor detection task to better utilize the wisdom of crowds and circumvent the sequence length limit.

\item Currently, the cross-lingual and cross-domain benchmarks for zero-shot rumor detection on social media still lack normative completeness, since the research on zero-shot rumor detection with propagation structure has just begun. The volume of low-resource languages like Arabic and even dialects is relatively hard to be organized with propagation structures on social media. Although this work collects a small annotated Arabic dataset corresponding to COVID-19 with propagation threads for model evaluation, we plan to evaluate our model on the datasets about more breaking events in low-resource domains and/or languages (e.g., Hindi) by leveraging existing datasets with rich annotation. Although there is a long way to go, where there is a will, there is a way.

\end{itemize}

\end{document}